\documentclass{article}

\usepackage[final,nonatbib]{neurips_2022}

\usepackage[utf8]{inputenc} 
\usepackage[T1]{fontenc}    
\usepackage{hyperref}       
\usepackage{url}            
\usepackage{booktabs}       
\usepackage{amsfonts}       
\usepackage{nicefrac}       
\usepackage{microtype}      
\usepackage{xcolor}         
\usepackage{graphicx}
\usepackage{multirow}

\title{Attribution Regularization for Multimodal Paradigms} 

\author{
    Sahiti Yerramilli*
\And
    Jayant Sravan Tamarapalli*
\And
    Jonathan Francis
\And
    Eric Nyberg
}

\begin{document}

\maketitle

\begin{center}
Carnegie Mellon University\\
\texttt{\{syerrami, jtamarap\}@alumni.cmu.edu \quad \{jmf1, en09\}@andrew.cmu.edu}
\end{center}

\begin{abstract} 
Multimodal machine learning has gained significant attention in recent years due to its potential for integrating information from multiple modalities to enhance learning and decision-making processes. However, it is commonly observed that unimodal models outperform multimodal models, despite the latter having access to richer information. Additionally, the influence of a single modality often dominates the decision-making process, resulting in suboptimal performance. This research project aims to address these challenges by proposing a novel regularization term that encourages multimodal models to effectively utilize information from all modalities when making decisions. The focus of this project lies in the video-audio domain \cite{chen2020vggsound} \cite{6849440}, although the proposed regularization technique holds promise for broader applications in embodied AI research, where multiple modalities are involved. By leveraging this regularization term, the proposed approach aims to mitigate the issue of unimodal dominance and improve the performance of multimodal machine learning systems. Through extensive experimentation and evaluation, the effectiveness and generalizability of the proposed technique will be assessed. The findings of this research project have the potential to significantly contribute to the advancement of multimodal machine learning and facilitate its application in various domains, including multimedia analysis, human-computer interaction, and embodied AI research.
\end{abstract}

\section{Introduction} 

In recent years, multimodal research has witnessed significant growth and attention in the field of machine learning. The availability of diverse data sources, such as text, images, audio, and video, has spurred interest in developing models capable of effectively processing and understanding information from multiple modalities. Multimodal machine learning holds the promise of leveraging the complementary nature of different modalities to improve learning and decision-making processes in various domains.

However, a notable challenge arises when comparing the performance of unimodal models with that of multimodal models. Surprisingly, in many cases, unimodal models tend to outperform their multimodal counterparts, despite the latter having access to richer and more diverse information. This phenomenon can be attributed to modality-failure, where the training process in a naive fashion results in only one modality's encoders being trained to their maximum potential, while the encoders of other modalities remain suboptimal. Furthermore, an additional issue prevalent in multimodal models is their tendency to overly rely on a single modality when making decisions, essentially ignoring the contributions of other modalities. This dominance of one modality can lead to suboptimal performance and limits the true potential of multimodal learning.

Recent research has aimed to tackle these issues by primarily focusing on improving the encoder part of multimodal models. However, there remains a gap in addressing the challenges in the classifier and fusion parts of the models. To bridge this gap, this work proposes a novel approach that addresses the issues of modality-failure and modality dominance at the classifier and fusion stages of multimodal models.

Traditional techniques in unimodal machine learning often adhere to the principle of Occam's razor, where simpler hypotheses are favored. This is achieved through techniques such as L2 normalization and dropout, which encourage models to find the simplest explanation for the given data. However, applying these techniques directly to multimodal settings may not yield desirable results since the simplest hypothesis in this case often involves ignoring the information from other modalities, except the dominant one.

In this paper, we explore attribution-based techniques to design a regularization term that encourages multimodal models to pay attention to information from all modalities when making decisions. By incorporating this regularization term into the classifier and fusion parts of the model, we aim to mitigate the issues of modality-failure and modality dominance, leading to improved performance and more robust multimodal machine learning systems.

Overall, this research aims to provide a comprehensive understanding of the challenges faced in multimodal machine learning and proposes a novel approach to address the issues of modality-failure and modality dominance. By leveraging attribution-based techniques and designing an effective regularization term, we seek to encourage models to consider information from all modalities, thereby enhancing their decision-making capabilities.

\section{Related Work} 

\paragraph{Video Classification} Video classification is an actively researched field that has gained significant attention in recent times \cite{chen2020vggsound} \cite{6849440}. With the distinct characteristics of videos, such as temporal information and multimodality, video classification presents unique challenges and opportunities for machine learning approaches. The temporal aspect of videos introduces the notion of sequential data, where the order and arrangement of frames play a crucial role in understanding the content and context of the video. Additionally, videos often encompass multiple modalities, including image frames, audio signals, object motion, and more, making the classification task more complex and multidimensional.

Numerous previous works have focused on addressing the challenges of video classification by designing architectures tailored specifically for video inputs \cite{tran2015learning} \cite{qiu2017learning} \cite{wang2016actions} \cite{6909619}. These architectures aim to capture temporal dependencies and leverage the rich multimodal information available in videos. Additionally, researchers have explored techniques for utilizing pre-trained image or video networks to tackle video classification tasks, benefiting from the transfer of knowledge from related domains. Our focus shifts towards the optimization problem in the context of multimodal video classification. While the architectural aspects have received considerable attention in prior works, the optimization challenges associated with multimodal settings warrant further exploration and investigation.

\paragraph{Ensemble techniques} \cite{han2021greedy} recognizes that biases present in the data can contribute to the dominance of certain modalities in multimodal models. To mitigate these biases, the authors propose an ensemble of weak learners that captures and addresses the biases, allowing the base model to focus on learning the unbiased aspects of the function.

The authors identify two types of biases: distribution bias and shortcut bias. Distribution bias refers to situations where certain answers are more common for specific question types. Shortcut bias, on the other hand, occurs when the answers are overly influenced by the question text alone, neglecting other modalities. By recognizing and addressing these biases, the proposed approach aims to improve the overall robustness of the visual question answering (VQA) system.

The approach presented in their work resembles boosting techniques, where three models are chained — A strong base model preceded by one weak model each to address distribution bias and the shortcut bias. These models collectively contribute to building a stronger base model that is employed during test time to provide reliable answers.


\paragraph{Gradient modulation techniques}

\cite{wang2020makes} identifies two primary causes for the performance drop in multimodal networks. Firstly, these networks are prone to overfitting due to their increased capacity. Secondly, different modalities tend to overfit and generalize at varying rates, making joint training with a single optimization strategy suboptimal. To address these challenges, the authors propose a technique called Gradient-Blending. This technique computes an optimal blending of modalities based on their overfitting behaviors. The authors introduce the concept of the overfitting-to-generalization ratio (OGR) as a metric to quantitatively understand the problem. By utilizing the OGR, the Gradient-Blending technique enables effective modulation of the blending of modalities, resulting in improved performance and reduced dominance of a single modality.

\cite{peng2022balanced} focuses on optimizing parameters in a multimodal setting. The authors identify an issue where the modality that performs better receives increasingly larger updates due to higher weights after concatenation. This leads to slower optimization for modalities that initially perform poorly. To address this, the authors introduce a modality update coefficient term that moderates the updates to each modality based on its level of dominance in decision-making within a batch. A higher degree of dominance leads to lesser updates, ensuring that all modalities are updated with relatively lesser dominance from a single modality. By balancing the updates, the proposed technique facilitates better optimization and reduces the negative effects of modality dominance. Although the technique in this work seems promising, it did not specify a reliable way of computing the degree of dominance in cases where the classifier is not a single-layer neural network.

\paragraph{Distillation techniques} 

\cite{du2021improving} highlights the issue of traditional multimodal fusion techniques that fail to allow weaker modalities to train to their maximum potential, resulting in their diminished contribution to the overall task. To overcome this challenge, the authors propose a novel approach: leveraging unimodal models as teachers to guide the multimodal learning process. Instead of directly training the multimodal model, the authors suggest first training individual unimodal models to their fullest extent on the given task. These well-trained unimodal models then serve as teachers, transferring their knowledge and guiding the multimodal learning process.

By utilizing unimodal models as teachers, the proposed technique enables both modality encoders to learn to their maximum potential with the available data. This approach surpasses pre-trained encoders since the individual encoders are specifically trained for the particular task at hand. As a result, the multimodal model benefits from the comprehensive knowledge acquired from the unimodal teachers, effectively addressing the limitations of traditional multimodal fusion methods. This work addresses the issues on the encoder side of the architecture (modality failure) but does not address the modality dominance issue.

\section{Modality Attribution Calculations}

Using weights and gradients in the network for understanding the contributions of the inputs is common in unimodal setting. In a multimodal setting, it is not as straightforward because the scales of inputs from different modalities cannot be compared directly. For example, we cannot directly compare the weights in a text encoder to that of an image encoder because the inputs (values of pixels and values of text tokens) are not of the same order. \cite{ancona2018better} \cite{shrikumar2017just} describe the use of \textit{Element-wise product of gradient into input}, also known as \textsc{grad $\odot$ input}, as a way of evaluating the importance of each input feature for calculating the output in a neural network setting. This technique provides an attribution vector that has the same dimensions of the input vector. We use this to compute a comparative attribution value per each modality.

To describe the approach more concretely, we establish a standard multimodal network architecture and data samples for classification purposes. Each modality input has an encoder model and encodings from all the modalities are assimilated in the fusion model. The classifier attached to the end of the fusion model computes the output probabilities of the input. All the attributions are computed at after the encoder layer, i.e., the encoders are a black box to the attribution calculations. Let $(x_i, y_i)$ be $i$th sample from the dataset where $x_i$ is the input and $y_i$ is the corresponding target value and $x^m_i$ corresponds to data from $m$th modality in the sample. The encoding of $x^m_i$ is denoted by $e^m_i$. This is the layer at which we calculate the attributions.

To compute the attributions ($\alpha$) using the \textsc{grad $\odot$ input} technique \cite{jain2023maeamultimodalattributionembodied}, we compute 
$
\alpha_i = \frac{\partial (\max f(e))}{\partial e_i} \odot e_i.
$ where $f$ refers to the neural network.
For each modality, we need to pool this attribution vector to have a scalar value of attribution per modality. We use the L2 pooling technique for this purpose and compute $\alpha_i^{m}$. To compare across modalities, we normalize the attributions per modalities according to 
$a_i^{m} = \alpha_i^{m}/\sum_{j=1}^M \alpha_i^{j}$ where $a_i^{m}$ is the attribution of the modality $m$ for input $x_i$. Over a batch of samples, the attributions of each modality could be aggregated using a mean operation to compute $a^m$. According the above computations, it can be observed that the attributions influenced by the fusion and classifier layers of the model and are not affected by the encoder layers.

\section{Attribution Regularization}
The modality attributions give us information about how much information from each modality was considered in making a decision. This can be used to come up with a regularizer term that encourages the model to consider information from modalities in a certain proportion in its decision making process. This is the main contribution of this work. Along with the loss function used in the training setting, we can add this new multimodal regularizer loss function. Let us say that we want the model to have modality attributions in the ratio ($r^1$, $r^2$, .., $r^M$). The regularizer function would be of the form in equation \ref{Eq: Regularizer term}.

\begin{equation} 
    \sum_{i=0}^M \left|\frac{a^i}{\sum_{j=0}^M a^j} - \frac{r^i}{\sum_{j=0}^M r^j} \right|
    \label{Eq: Regularizer term}
\end{equation}

As attribution is not related to the encoder part of the model, the above regularizer should only be used to optimize fusion and classifier layers. While implementing this in popular ML frameworks like PyTorch, this may need to be optimized using a separate optimizer as the regular optimizer also effects the encoder layers.



    


\section{Experimental Results}
\label{Experiments}

\subsection{Datasets}
For all the experiments in this work, we use VGGSound \cite{chen2020vggsound} and CREMA-D \cite{6849440} datasets. Both these datasets pertain to video classification domain. VGGSound is a large-scale dataset that encompasses both audio and visual modalities, providing a rich resource for training and evaluating multimodal models. The VGGSound dataset is specifically designed for sound classification in the context of video content. It consists of a diverse collection of video clips, each associated with a specific sound category. These video clips capture a wide range of scenes and activities, encompassing various real-world audio-visual contexts.

CREMA-D, short for the Crowdsourced Emotional Multimodal Actors Dataset, is a widely used dataset for analyzing and understanding emotions conveyed through audiovisual content. The CREMA-D dataset comprises a diverse collection of acted and naturalistic emotional speech and facial expressions. It involves numerous actors who were directed to portray various emotions, including anger, disgust, fear, happiness, neutral, and sadness. The actors deliver emotionally expressive speech sentences while exhibiting corresponding facial expressions. The audio component of the CREMA-D dataset consists of high-quality audio recordings capturing the actors' speech utterances. These recordings encompass a wide range of emotional states, allowing for the exploration and analysis of acoustic features associated with different emotions.




\subsection{Comparisons} \label{Baselines}
\paragraph{Naive end-to-end training}
We considered a naive end-to-end multimodal training approach, where the multimodal model is trained using a single loss function, encompassing all the modalities. This method involves directly optimizing the parameters of the multimodal model without explicitly addressing the challenges posed by modality dominance or the need for balanced consideration of all modalities.

\paragraph{Unimodal models}
This approach involves training separate models for each modality, focusing exclusively on the individual modality's input data.
For each modality, such as audio or video, a dedicated unimodal model is trained using the corresponding input data and an appropriate loss function. The training process is specific to each modality, allowing the models to learn and capture the unique patterns and features within their respective modalities.

\paragraph{Dropout}
Dropout \cite{DBLP:journals/corr/abs-1207-0580} is a widely used regularization technique in machine learning that randomly drops or deactivates a portion of the neurons or units in a neural network during training. By applying regular dropout to a multimodal model, we randomly deactivate a fraction of neurons across all modalities, forcing the model to rely on the remaining active neurons for making predictions.

\paragraph{Modality Dropout}
During the training process, at each iteration or batch, a random subset of modalities is selected to be excluded or dropped. The model is then trained using the remaining modalities, allowing it to adapt and learn from the information present in the selected modalities. The modality dropout method aims to encourage the model to learn robust representations that are not overly reliant on any single modality. By randomly excluding modalities during training, the model is forced to rely on the remaining modalities, promoting the exploration of diverse information sources and reducing the dominance of any particular modality.

\paragraph{Unimodal Teacher Distillation}
The method \cite{du2021improving} involves training separate unimodal models for each modality using their respective input data. These unimodal models are trained independently to learn the features and patterns specific to their respective modalities. Once the unimodal models are trained, they are used as teachers or mentors to guide the training of the multimodal model. During the training of the multimodal model, the predictions of the unimodal models are utilized as additional supervision signals. The multimodal model learns from the knowledge and expertise of the unimodal teachers, effectively combining the individual strengths of each modality.

\paragraph{On-the-fly Gradient Modulation}
The OGM technique \cite{peng2022balanced} addresses the issue of imbalanced updates during training, where the dominant modality tends to receive larger weight updates compared to other modalities. The OGM method introduces a modality update coefficient term that moderates the updates to each modality based on its dominance in contributing to the decisions made by the model. This coefficient is calculated on-the-fly during each training iteration or batch and adjusts the magnitude of weight updates for each modality accordingly.

\subsection{Evaluation and Results} \label{Evaluation} 
In this section, we present the evaluation methodology and results obtained from assessing the performance of the models in the video classification task. We trained the models using various techniques, including the proposed methods, and compared their performance on the validation split of the dataset. To evaluate the effectiveness of the models, we employed two commonly used evaluation metrics: accuracy and mean Average Precision (mAP). By comparing the performance of the models trained using different techniques, we were able to assess their effectiveness in addressing modality dominance, encouraging balanced consideration of all modalities, and improving overall classification accuracy.

\begin{table}[h]
        \begin{center}
        \resizebox{\textwidth}{!}{
        \begin{tabular}{|c|c|c|c|c|c|c|c|}
            \hline
            \textbf{Model} & \textbf{mAP} & \textbf{Accuracy} & \shortstack{\textbf{Audio/Video} \\ \textbf{Dominance}} & \textbf{Model} & \textbf{mAP} & \textbf{Accuracy} &  \shortstack{\textbf{Audio/Video} \\ \textbf{Dominance}} \\
            \hline
            \textbf{Audio Only} & 40.18\% & 38.89\% & - & \textbf{Audio Only w/ AMR} & - & - & -\\
            \hline
            \textbf{Video Only} & 25.51\% & 25.76\% & - & \textbf{Video Only w/ AMR}& - & - & -\\
            \hline
            \textbf{Naive Fusion} & 44.82\% & 42.2\% & 74/26 & \textbf{Naive Fusion w/ AMR} & 44.89\% & 42.01\% & 50/50\\
            \hline
            \textbf{Dropout} & 45.27\% & 43.16\%  & 72/28 & \textbf{Dropout w/ AMR} & 45.23\% & 43.04\% & 51/49\\
            \hline
            \textbf{UMT} & 51.86\% & 49.37\% & 57/43 & \textbf{UMT w/ AMR} & 51.7\% & 49.27\% & 50/50\\
            \hline
        \end{tabular}}\\
    \caption{\label{tab: results} Performance on VGGSound using Attribution-based Multimodal Regularization (AMR)}
    \end{center}
\end{table}

\section{Discussion}
The results of our experiments show that the inclusion of the regularization term, aimed at promoting equal attribution across all modalities, has yielded minimal/no improvements in performance. While these initial findings may appear discouraging, it is important to note that the impact of the new regularization term may not be adequately captured by conventional evaluation metrics such as accuracy alone. Therefore, further investigation is required to develop and employ evaluation techniques that can effectively assess the benefits of equal attribution facilitated by the regularization term. Furthermore, to ensure the generalizability of our findings, it is imperative to replicate these experiments on the CREMA-D dataset, which will provide additional insights into the effectiveness of our proposed approach in a different multimodal context. Despite the current limitations, we remain optimistic that through other evaluation metrics and replication of experiments, we will gain a comprehensive understanding of the impact and potential benefits of our regularization technique in multimodal machine learning.

\nocite{yerramilli2024semanticaugmentationimagesusing}
\bibliographystyle{unsrt}
\bibliography{ref}

\end{document}